\documentclass[lettersize,journal]{IEEEtran}
\usepackage{amsmath,amsfonts}
\usepackage{algorithmic}
\usepackage{algorithm}
\usepackage{array}
\usepackage[caption=false,font=normalsize,labelfont=sf,textfont=sf]{subfig}
\usepackage{textcomp}
\usepackage{stfloats}
\usepackage{url}
\usepackage{verbatim}
\usepackage{graphicx}
\usepackage{cite}
\usepackage{multirow}
\usepackage{authblk}
\begin{document}
\title{Analysis of Data Augmentation Methods \\for Low-Resource Maltese ASR}


\author{Andrea DeMarco,
Carlos Mena,
Albert Gatt,
Claudia Borg,
Aiden Williams, and
Lonneke van der Plas
\thanks{This paper was submitted for review on the 12th October, 2022. This research was supported in part by a University of Malta Research Excellent Grant.
\par{A. DeMarco is with the Institute of Linguistics and Language Technology, University of Malta (e-mail: andrea.demarco@um.edu.mt).}
\par{C. Mena was with the Institute of Linguistics and Language Technology, University of Malta, and is now with the Language \& Voice Lab, Reykjavik University (e-mail: carlosm@ru.is)}.
\par{A. Gatt was with the Institute of Linguistics and Language Technology, University of Malta, and is now with the Department of Information and Computing Sciences, Utrecht University, The Netherlands (e-mail: a.gatt@uu.nl).}
\par{C. Borg is with the Department of Artificial Intelligence, University of Malta (e-mail: claudia.borg@um.edu.mt).}
\par{A. Williams was a student at the Department of Artificial Intelligence, University of Malta (e-mail: aiden.williams.19@um.edu.mt).}
\par{L. van der Plas was formerly at the Institute of Linguistics and Language Technology, University of Malta, and is now with the Idiap Research Institute, Switzerland (e-mail: lonneke.vanderplas@idiap.ch).}}
}



\maketitle

\begin{abstract}
Recent years have seen an increased interest in the computational speech processing of Maltese, but resources remain sparse. In this paper, we consider data augmentation techniques for improving speech recognition for low-resource languages, focusing on Maltese as a test case. We consider three different types of data augmentation: unsupervised training, multilingual training and the use of synthesized speech as training data. The goal is to determine which of these techniques, or combination of them, is the most effective to improve speech recognition for languages where the starting point is a small corpus of approximately 7 hours of transcribed speech. Our results show that combining the data augmentation techniques studied here lead us to an absolute WER improvement of $15$\% without the use of a language model.
\end{abstract}


\section{Introduction}
\label{sec:intro}
\IEEEPARstart{I}{n} recent years, Automatic Speech Recognition (ASR) has seen heavy interest in large-scale reusable multilingual models~\cite{Baevski2020}, relying on large quantities of pre-training data. There has been a lot of work focused on transfer learning-based improvements to ASR systems in order to make many modern architectures accessible to a wider set of languages. For example~\cite{wang2020improving} use high-to-low resource language machine translation as an intermediate task. Recent research has also turned towards large, multilingual, pretrained ASR architectures. For example~\cite{pratap2020} show that great benefits for low-resource speech recognition can accrue from a multilingual acoustic model. However, the majority of languages used in this work have over 100 hours of training data available. A different line of research has been pursued in Wav2Vec2~\cite{baevski2020wav2vec}, which builds on the earlier Wav2Vec model~\cite{Schneider2019}. This architecture can learn robust speech representations of sufficient quality to perform recognition even for target languages with a very low volume of labelled data (in the order of one hour), reaching the same performance as previous state-of-the-art systems requiring 100 times more data (100 hours). Wav2Vec2 can also provide good results with just ten minutes of labelled speech data given substantial pretraining on 53,000 hours of unlabelled data. This work therefore demonstrates that semi- or self-supervised pretraining can result in robust representations that lower the requirements on labelled data. Once again, however, the key is in ensuring that sufficient data is available for pretraining, in this case in the form of audio. Indeed, \cite{baevski2020wav2vec} describe two settings, one with 960 hours of unlabelled data, and another with 60,000 hours.
\IEEEpubidadjcol
The main selling point of pretraining-based approaches is that they make highly effective use of large, unlabelled datasets. There are however two challenges with ensuring that there is sufficient data for a particular language. The first is quality: unfiltered data sourced from the web can be extremely noisy and of poor quality, with a high incidence of automatically translated sources, among other problems~\cite{kreutzer_quality_2022}. The second is that a much lower quantity of data is available for under-represented languages, compared to, say, English or Mandarin. This leads to an impoverished resource landscape for such languages~\cite{besacier2014automatic} and contributes to the lack of representation of demographic or ethnic groups in AI~\cite{Bender2021}. This quantity challenge is exacerbated if web-sourced data is also filtered to address quality. To take an example, in the 2022 release of OSCAR~\cite{abadji_towards_2022}, a multilingual dataset extracted from CommonCrawl and passed through a quality control pipeline, over 50\% of the languages have less than 11m tokens. One of these languages is Maltese\footnote{The OSCAR Maltese dataset consists of ca 118k tokens, in 2200 documents. For comparison, the Maltese version of Wikipedia, a frequently used source of textual data for multilingual models such as mBERT~\cite{Devlin2019}, consists at the time of writing of 2214 documents.}. In a recent study of multilinguality in Natural Langauge Processing, Maltese was classified (with other languages such as Irish and Zulu) as being among those languages for which labelled data is also sparse, though the situation is improving~\cite{joshi_state_2020}. Thus, languages such as Maltese provide a good test-case for when web-scale textual or speech data is harder to obtain.

This paper presents an exploratory analysis of a set of related techniques for improving ASR for Maltese using different data augmentation strategies, relying on the use of smaller, curated datasets in the target language, the use of larger datasets in related languages, and the contribution of artificial data generation. Briefly, the scenario in which these experiments were designed consists of the following: (i) approximately 7 hours of labelled Maltese (target language) speech data~\cite{mena2020masri}, (ii) a small amount of unlabelled speech data in the target language, (iii) a medium sized corpus of Maltese text (250m tokens)~\cite{gatt2013,camilleri2013}, (iv) a diphone/concatenative TTS system whose output, while not of high quality, can be exploited for data augmentation purposes~\cite{Borg:14}, (v) substantial data in a number of other languages -- notably Arabic, Italian and English -- which are typologically and historically related to Maltese~\cite{brincat2011,borg2017morphological}.


\section{Typological relation of Maltese with other languages}
\label{sec:intro2}
Maltese bears strong historical relationships to three major languages. From a historical perspective, it has been characterised as having a Semitic/Arabic stratum, a Romance (Italian/Sicilian) superstratum, and an English adstratum~\cite{brincat2011}. 
Originally a variety of Arabic, Maltese came into intensive contact with Italian due to its geographical proximity and historical relations with that country. This was followed by a period of intensive contact with English (Malta was a British colony from 1800 to 1964), as a result of which, the Maltese population is largely bilingual. The impact of this linguistic history is clearly evident in the language at many levels of analysis, including the lexical and morphological~\cite{borg2017morphological}. From the perspective of the present experiments, this suggests that a promising way to approach the data augmentation problem is to exploit the (much larger) resources available for these languages. We do this by transcribing speech data from one of these languages with a baseline Maltese acoustic model trained on a very small corpus (Section \ref{subsec:unsup-cycles}); and using a `language mixture' approach (Section \ref{subsec:lang-mixes}), in which an end-to-end system is pretrained on data using mixtures of such languages. In both cases, pretraining is followed by a fine-tuning step.


\section{Data augmentation methods}
\label{subsec:intro-review}

Data augmentation is a technique utilized in the field of machine learning that consists of modifying the existing data in some way, or using data from other sources that are easy and/or cheap to obtain, in order to increase the effective size of the corpus with which a system is trained, obtaining better performance. 
For example, one of the 
earliest applications of data augmentation was applied in the field of image processing. It consists in rotating the images of the dataset multiple times~\cite{simard2003best}, in order to increase the amount of the effective images in the training set.

Several types of data augmentation methods are used in 
speech recognition. In this section, we present a review of these methods, from among which we identify those that
fit better with the small amount of Maltese language data available and with modest computational resources.

According to the taxonomy presented in~\cite{ramirez2019survey}, the types of data augmentation for ASR can be divided in three groups: semi-supervised training, transformation of acoustic data, and speech synthesis. On the other hand, in~\cite{ragni2014data} the authors argue that ``the schemes can be split based on the type of produced data, such as unsupervised, synthesized and other language data". In~\cite{cui2015data}, it is mentioned that there exists one kind of data augmentation that uses ``label-preserving transformations"~\cite{jaitly2013vocal}, where data is altered but not its labels. This latter type of data augmentation can include several subtypes, such as transformation of acoustic data and other-language data - this however does not qualify as unsupervised or semi-supervised training, because in such cases, the labels are generated, and not ``preserved" from their original form.

One other issue that comes out of the literature is the different definitions  assigned to the different methods of data augmentation; for example in~\cite{ragni2014data}, the authors argue that ``the unsupervised data refers to data which lacks correct transcriptions. This also includes data having only rough transcriptions, such as closed captions". This definition is problematic because it overlaps with the definition of semi-supervised training given in~\cite{ramirez2019survey}, which claims: ``the semi-supervised training approach assumes the use of the text produced by an automatic speech recognition system to train acoustic models". The problem is that the transcriptions produced by an ASR system can also be full of errors; more than that, closed captions can also be generated by an ASR system.

Given the somewhat variable nature of terminology and classification in the field, in the remainder of this section, we organise related work using our own data augmentation taxonomy. Our goal is to preserve, as much as possible, the main ideas found in literature, but at the same time, avoid clashes between definitions. Our taxonomy is as follows:
\begin{enumerate}
    \item Semi-supervised training
    \item Data perturbation
    \item Training with synthesized data
    \item Unsupservised training
    \item Multi-lingual training
\end{enumerate}

\subsection{Semi-supervised training}
\label{subsec:semi-sup-tra}
Semi-supervised data refers to data which lacks ``gold" transcriptions~\cite{ragni2014data}. These transcriptions can be obtained in different ways. For example, they can be supplied manually by non-expert transcribers, or by harvesting closed captions. Semi-supervised training is sometimes also referred to in the literature as ``lightly-supervised training''~\cite{gales2006progress}. Some examples of this can be found in~\cite{ragni2014data, gales2006progress, lamel2002lightly}.

\subsection{Data perturbation}
In this approach, the speech signal is modified in some way, but assigned the same associated transcription (label). The most common example of this scheme is to add noise to the speech signal~\cite{ramirez2019survey, jimerson2018improving}. Variation of the speech rate and the pitch~\cite{thai2019synthetic, vachhani2018data, kanda2013elastic} is also a popular technique. Vocal Tract Length Perturbation (VTLP) is a more sophisticated method to perturb the input signal and it is part of this category as well~\cite{billa2018isi, jaitly2013vocal, cui2015data}. More recently, SpecAugment~\cite{park2019specaugment} was proposed, performing rectangular cuts in the log mel spectrogram. We integrated SpecAugment as part of our experiments.

\subsection{Training with synthesized data}
Data augmentation based on synthesized data refers to new, artificially generated data. For example, it is common practice to create synthesized speech from raw text with a Text to Speech (TTS) system, and then combine it with a regular ASR corpus~\cite{rossenbach2020generating, tjandra2017listening, rosenberg2019speech}. Nevertheless, one can find more sophisticated types of synthesized data, as in~\cite{gales2009support}, where some Hidden Markov Models (HMMs) are created artificially to help support vector machines (SVMs) to outperform their classification capabilities. In our experiments, we use synthesized speech from a simple diphone concatenative TTS system~\cite{Borg:14}. No corpus of speech for this TTS system was utilized - and building such a TTS system lends well to a low-resource language such as Maltese.

\subsection{Unsupervised training}
Unsupervised training data refers to data with transcriptions generated only by an ASR system. It can be confused with the semi-supervised approach in the sense that the unsupervised transcriptions can have a lot of errors too, but as discussed in 
sections~\ref{subsec:intro-review} and ~\ref{subsec:semi-sup-tra}, the semi-supervised transcriptions can come from a human source, and not necessarily from an ASR system. That is the main difference. 

In this paper, we distinguish unsupervised training from the semi-supervised training discussed in section~\ref{subsec:semi-sup-tra}. For convenience, we reserve the term ``semi-supervised'' to noisy transcriptions obtained from human annotators. The term ``unsupervised'' training will refer only to setups involving data from transcriptions generated by an ASR system.

Unsupervised training is commonly associated in the literature with transcribing speech in a language A, using an ASR system trained in a language B, as in~\cite{ragni2014data, loof2009cross, qian2013combination}. However, this need not be the case. One can transcribe speech in a language A using an ASR system in the same language~\cite{zavaliagkos1998utilizing, cucu2014unsupervised, kemp1999unsupervised,wessel2004unsupervised}. In our experiments, we use an ASR system in Maltese to decode other languages as well as our target language (see sections~\ref{subsec:unsup-cycles},~\ref{subsec:lang-mixes},~\ref{subsec:synthetic}). 

\subsection{Multilingual training}
In this scheme of data augmentation, the idea is to take advantage of the speech in a language different from the target, in order to augment the training data available for the desired system. In the literature, related procedures are sometimes referred to as ``data borrowing''~\cite{qian2013combination} or even ``out-of-language data''~\cite{imseng2014using}. 

However, the way in which this multilingual data is used may vary, depending on factors such as the amount of data available in the target language or the type of system utilized to do the training. For example, a common scenario is when the amount of target language data is small, but then it is combined with data in other languages~\cite{billa2018isi, yilmaz2018acoustic, qian2013combination, imseng2014using}. Nevertheless, the scenario of not having any data at all in the target language can also occur, as in~\cite{loof2009cross}.

Sometimes the original transcriptions in the other, non-target languages are available, so one way to combine all the languages in one system is using a phoneme set in common~\cite{billa2018isi, loof2009cross, knill2013investigation, vu2013multilingual, grezl2014adaptation}. Sometimes, for various reasons, transcriptions in the non-target 
languages are not available or they are not usable; in that case, one can use an ASR system in the target language to produce unsupervised transcriptions of the other languages, in order to use them to train an ``augmented" acoustic model that will be the combination of the data in the target language plus the data in other languages~\cite{ragni2014data,qian2013combination}.

Commonly used techniques for developing such multilingual systems include multi-layer perceptrons (MLP)~\cite{tuske2014data, qian2013combination, vu2013multilingual, imseng2014using} and bottleneck features~\cite{tuske2014data, knill2013investigation, tuske2013investigation}.

In our particular case, we will be using an End-to-End (E2E) system to produce unsupervised transcriptions from distinct languages which are typologically or historically related to the target language Maltese, such as Arabic, Italian and English (see sections~\ref{subsec:unsup-cycles},~\ref{subsec:lang-mixes},~\ref{subsec:synthetic}).In addition to that, we experiment with the original transcriptions of some languages related to Maltese.

\section{Data and resources}
\label{sec:data}
In order to explore two principal data augmentation strategies, namely (i) creating synthetic data, and (ii) using both semi-supervised/unsupervised (noisy) and supervised (gold) data from related languages in a transfer setting, a number of tools and datasets were utilized.

\subsection{Speech synthesis}
\label{subsec:tts-system}
The Maltese Speech Engine\footnote{The Maltese Speech Engine is available for download {\url{https://fitamalta.eu/projects/maltese-speech-engine-synthesis-erdf-114/download-maltese-speech-engine/}} and runs on MS Windows.}~\cite{Borg:14} is a diphone-based synthesis system. It relies on a rule-based grapheme to phoneme (G2P) tool, which transforms text into its equivalent IPA representation. Since Maltese is fairly homographic, the rules are primarily based on one-to-one mappings with a list of exceptions.\footnote{A reimplemented Python version of the G2P tool is available from the Maltese Language Resource Server: \url{https://mlrs.research.um.edu.mt/index.php?page=api}. A demo is also available on the same server.}

Although the TTS speech engine produces mostly intelligible Maltese, it is far from natural-sounding and contains several pronunciation errors. This justifies our labelling the data created using this engine as `noisy'. Nevertheless, we contend that such noisy data can still prove useful in pretraining, if followed by fine-tuning on carefully curated speech-text pairs. Hence, we used the Maltese Speech Engine to create a synthetic corpus, described in section~ \ref{subsec-masri-synth}.

\subsection{Korpus Malti text corpus}
\label{subsec:mlrs-corpus}
The Korpus Malti v3.0~\cite{gatt2013} is a text corpus of around 250m tokens in several different genres, including parliamentary debates, news, law, opinion articles, sports articles, culture, academic, literature and religious texts.\footnote{The corpus is available on the Maltese Language Resource Server, and can also be searched through an online interface. See: \url{https://mlrs.research.um.edu.mt/index.php?page=corpora}}. Tokens in the corpus are tagged with part of speech, and labelled with lemmas and (where relevant) consonantal root for words of a Semitic origin. We use this text corpus for creating synthesized data as described in section~\ref{subsec-masri-synth}.

\subsection{MASRI-Synthetic Corpus}
\label{subsec-masri-synth}
We produced a synthetic speech corpus using the Maltese Speech Engine described above, by feeding the speech engine sentences from the Korpus Malti v3.0. A total of 52,500 sentences were sampled, after filtering according to the following criteria:

\begin{itemize}
    \item All sentences sampled consisted of 30 words or less;
    \item No sentences containing non-Maltese orthographic characters (such as {\em y}, which only occurs in loan words), foreign words or proper names were included. 
\end{itemize}

The resulting speech corpus, which we refer to as the MASRI-SYNTHETIC corpus, consists of $105$ male and $105$ female voices, produced with varied pitch (21 values of pitch ranging between $-20$ and $20$) and speech rate (5 values between $-2$ and $2$). Each voice was assigned 250 utterances of 13 words each, resulting in utterances of varying length between 2 and 10 seconds duration. The corpus has a total duration of 99 hours and 18 minutes.

\subsection{The MASRI-HEADSET corpus}
\label{subsec:masri-headset}
The MASRI-HEADSET Corpus~\cite{mena2020masri} is, to our knowledge, the first published ASR corpus in Maltese. It comprises $8$ hours of read speech, recorded in a low-noise environment. The corpus contains $25$ different speakers ($13$ female; $12$ male) and includes orthographic transcriptions. We refer to this corpus as the MASRI-HEADSET version $1$. In our experiments, we split the corpus into two unique portions: a training set, with a size of $7$ hours and $34$ minutes; and a test set of $31$ minutes. 

Since its publication, MASRI-HEADSET v1 was further modified to improve quality, by removing excessive periods of silence, suppressing clicks and segmenting audio files longer than 11 seconds. This resulted in what we refer to as MASRI-HEADSET version $2$, with a total duration of 06h39m, against the original 08h06m.

For the purposes of the present experiments, we used the same amount of utterances (250) in the test split of MASRI-HEADSET v2 as in the original v1 corpus. This resulted in a duration of $6$ hours and $19$ minutes for the training set and $19$ minutes for the test set.

\subsection{Maltese audiobooks}
\label{subsec:mal-abooks}
As an additional source of untranscribed data, we use a set of audiobook recordings, obtained from a Maltese publisher. These consist of 12 hours and 24 minutes of recorded speech. 

\subsection{Speech Corpora from Related Languages}
Another aspect of our research is to experiment with transfer learning from related languages. Thus we used a number of other corpora.

\subsubsection{LibriSpeech}
\label{subsec:librispeech}
LibriSpeech\footnote{LibriSpeech can be downloaded from: \url{http://www.openslr.org/12}} is an ASR corpus of read speech in English~\cite{panayotov2015librispeech}, derived from audiobooks that are part of the LibriVox project\footnote{LibriVox project allows volunteers to read and record chapters from books in the public domain. See: \url{https://librivox.org/}}. It contains around 1,000 hours of speech sampled at 16 kHz and partitioned into different training, development and test sets according to quality. For our work, we combined all the different training sets into a single pool, irrespective of the quality, and then randomly selected male-female pairs of speakers. The resulting selection from LibriSpeech is a 100-hour corpus with the only constraint that total speech time is perfectly balanced across the two genders. 

\subsubsection{Arabic Speech MGB-2 corpus}
\label{subsec:arab-data}
Given the historical relationship between Maltese and Arabic,
we include Arabic as part of our transfer learning experiments. Arabic Speech\footnote{\url{https://arabicspeech.org/}} is a collaborative effort aimed at increasing speech resources for Arabic and its dialects. For this research, we selected the MGB-2 corpus, a multi-dialect broadcast news corpus consisting of 1,200 hours, collected from Aljazeera TV programs and made available by the Arabic Speech project. This corpus was used in the 2016 MGB challenge~\cite{ali2016mgb}. For our purposes, we used a 100-hour portion.

This corpus includes transcriptions in both Arabic and the Buckwalter (BW) transliteration~\cite{habash2007arabic}. Since Maltese uses the Roman alphabet, and the models we experiment with rely on latin characters, we could not use Arabic transcriptions. We therefore rely on BW transliterations, noting that these lack vowel annotations, which is a source of potential ambiguity for an ASR system. Given the provenance of the Arabic data, it is worth noting that it is noisier than the other corpora described here. In particular, several audio files contain audible speech from more than one person (e.g. a newscaster, with a simultaneous translator), as well as background noise or music.

We use a 100-hour, random sample of the corpus. Due to the gender imbalance in the corpus as a whole (a far higher representation of male speakers), the sample is also not balanced for male versus female voices.

\subsubsection{Italian M-AILABS speech dataset}
\label{subsec:ita-data}
The M-AILABS Speech Dataset\footnote{\url{https://www.caito.de/2019/01/the-m-ailabs-speech-dataset/}} is an open source corpus combining data from LibriVox and Project Gutenberg\footnote{Project Gutenberg is a collection of books that are free from copyright. See: \url{http://www.gutenberg.org/}}. The data consists of nearly a thousand hours of audio clips of read speech in eight different languages. Clips vary in length, from 1 to 20 seconds, and include the corresponding transcriptions. In our experiments, we only use the Italian portion of this corpus with a total length of $118$ hours and $41$ minutes. This portion was taken as is, with no gender balancing or any other preprocessing made to the audio or text data.


\section{Methodology}
\label{sec:methodology}
In this section, we describe in general terms, how the experiments in Section~\ref{sec:experiments} are performed, as well the features that they have in common. All the experiments reported in section~\ref{subsec:unsup-cycles} onwards, are split into two separate stages: pretraining and fine-tuning. Data augmentation methods are applied during the pretraining stage. The fine-tuning stage uses a pretrained model as a starting point or bootstrap. Fine-tuning is always performed with the training portion of MASRI-Headset. All experiments utilise the MASRI-Headset test set.

\subsection{Architectures}
\label{subsec:arch}
In all our experiments, we use Jasper (Just Another Speech Recognizer), a family of End-to-End speech recognizers created by NVIDIA~\cite{li2019jasper}. We use the version included in the Neural Modules (NeMo) framework\footnote{\url{https://github.com/NVIDIA/NeMo}}~\cite{kuchaiev2019nemo}. 

A Jasper architecture typically consists of $n$ consecutive blocks, each with $r$ convolutional layers or sub-blocks. Each sub-block is composed of a 1D separable convolution followed by batch normalisation, a ReLU layer and dropout. Finally, all Jasper architectures have four additional convolutional blocks: one for pre-processing and three for post-processing. Table~\ref{tab:pre-proc-layer} summarises the pre-processing parameters used for all experiments.


\begin{table}[ht]
\caption{Jasper preprocesing layer parameters}
\label{tab:pre-proc-layer}
\centering
\begin{tabular}{lll}
\noalign{\smallskip}\hline\noalign{\smallskip}
\textbf{Parameter} & \textbf{Value} \\
\noalign{\smallskip}\hline\noalign{\smallskip}
Window type & Hanning\\
Window size & 20ms\\
Window stride & 10ms\\
Padding & 16 \\
Feature type & logbank\\
Normalization scope & Per feature\\
Number of features & 64\\
Apply Short-Time Fourier Transform (STFT)? & Yes\\
Fast Fourier Transform (FFT) data points & 512\\
Dither value & $0.00001$\\
\noalign{\smallskip}\hline\noalign{\smallskip}
\end{tabular}
\end{table}

A number of execution parameters are common to all experiments without exception; these are summarised in Table~\ref{tab:common_params}. As shown in the table, we use the Novograd optimizer, which is similar to Adam~\cite{kingma2014adam}, except that its second moments are computed per layer instead of per weight~\cite{li2019jasper}. The loss is calculated based on the Connectionist Temporal Classification (CTC) algorithm~\cite{graves2006connectionist}.

\begin{table}[ht]
\caption{Common execution parameters of all the experiments}
\label{tab:common_params}       
\centering
\begin{tabular}{lll}
\noalign{\smallskip}\hline\noalign{\smallskip}
\textbf{Parameter} & \textbf{Value} \\
\noalign{\smallskip}\hline\noalign{\smallskip}
Batch size & $32$ \\
Number of epochs & $50$ \\
Learning rate & $0.02$ \\
Optimizer & Novograd \\
Loss function & CTC \\
\noalign{\smallskip}\hline\noalign{\smallskip}
\end{tabular}
\end{table}

\subsection{Generation of noisy transcriptions}
A number of experiments rely on data augmentation using semi-supervised  and unsupervised (`noisy') transcription. This is always performed in the following way: a set of audio files in some language is transcribed using an acoustic model trained with the training portion of MASRI-Headset to produce transcriptions in Maltese. These transcriptions can have a real meaning when the input files are in Maltese, but if the input is in a different language, the transcriptions may only contain text that looks like Maltese, but it is not (`pseudo-Maltese'). It has to be emphasized that all the noisy transcriptions produced in the present work are only in Maltese or pseudo-Maltese.

\subsection{Noising via SpecAugment}
\label{subsec:use-specaugment}
In all experiments, we use SpecAugment~\cite{Park2019}, a technique which has been shown to improve the robustness of end-to-end ASR models, by augmenting data with noise introduced directly into the feature inputs of the model. Table~\ref{tab:specaugment-params} summarises the parameters used. In general, SpecAugment parameters for $100$ hours or more are applied in the pretraining stage, while parameters for less than $100$ hours are applied during fine-tuning. 

\begin{table}[ht]
\caption{SpecAugment parameters}
\label{tab:specaugment-params}
\centering
\begin{tabular}{lll}
\noalign{\smallskip}\hline\noalign{\smallskip}
\textbf{Parameter} & \textbf{$<100h$} & \textbf{$\geq 100h$} \\
\noalign{\smallskip}\hline\noalign{\smallskip}
Rectangles placed randomly & $5$ & $5$ \\
Vertical stripes (time) & $2$ &  $120$\\
Horizontal stripes (frequency) & $2$ & $50$\\
\noalign{\smallskip}\hline\noalign{\smallskip}
\end{tabular}
\end{table}

\subsection{Experimentation without language models}
\label{subsec:no-lm}
None of the experiments reported below incorporates language model (LM) rescoring. There are two reasons for this. First, once an LM is used to rescore and improve ASR output, it results in modifications of the transcriptions produced by an acoustic model. This occurs because an LM adjusts the transcriptions produced by the acoustic model, based on the LM vocabulary.\footnote{This is also why, in a classic ASR system based on HMM, both the vocabulary extracted from the LM and from the training transcriptions, have to be included in a pronunciation model such as a pronouncing dictionary.}~\cite{huang2001spoken,jurafsky1995using}. Since our goal is to determine the impact of noisy or synthetic data augmentation on acoustic model quality, the inclusion of LM rescoring would obscure the contribution of these techniques. Clearly, the results reported in the remainder of this paper could be improved by the inclusion of LM rescoring, but this is orthogonal to the present goals.

The second reason is a direct consequence of the small amount of Maltese data available. The only ASR corpus used at the time the present experiments were performed is MASRI-HEADSET (see section~\ref{subsec:masri-headset}), whose sentences were extracted from the Korpus Malti v3.0 (Section~\ref{subsec:mlrs-corpus}). The latter was also the largest available corpus on which a language model of reasonable size could be built. Rescoring with an LM built on this corpus would give better results for an acoustic model trained on MASRI-HEADSET (the baseline model described in Section~\ref{subsec:validation} below), given the substantial overlap in their vocabulary. On the other hand, filtering the test utterances from Korpus Malti v3.0 prior to creating the LM would introduce a bias against the test utterances of the MASRI-HEADSET which are sourced from  the same corpus. 

To demonstrate these potential biases, we created two different LMs using the settings described by~\cite{mena2020masri}. The Maltese text was extracted from the Korpus Malti v3.0 and excluded sentences containing extensive borrowing from English, as well as digits and proper names.\footnote{English words were identified using the CMU Pronounciation Dictionary~\cite{weide1998cmu}. To identify proper names we used the CIEMPIESS-PNPD~\cite{mena2019opencor}.} Our initial LM (which is biased in favour of the baseline) is a 6-gram model (ca. $9.5MB$) which includes test set sentences from MASRI-HEADSET. A second LM (`against') was created by filtering the test set sentences (ca. $8.7$MB). 

Table~\ref{tab:LM_Results} shows the results obtained on the MASRI-HEADSET test set, using acoustic models trained only with the training portion of the same corpus (the baselines described in Section~\ref{subsec:validation}), as well as the best model obtained in our experiments (Section~\ref{subsec:synthetic}). We show results both with and without rescoring using either of the two LMs.

\begin{table}[ht]
\caption{Word Error Rate (WER) for baseline and best models, with and without the LMs}
\label{tab:LM_Results}       
\centering
\begin{tabular}{lll}
\noalign{\smallskip}\hline\noalign{\smallskip}
 & \textbf{Baseline} & \textbf{Best}  \\
\noalign{\smallskip}\hline\noalign{\smallskip}
No LM & $63.71$\% & $48.97\%$ \\
LM Against & $59.04$\% & $54.67\%$ \\
LM in Favor & $20.87$\% & $16.65\%$ \\
\noalign{\smallskip}\hline\noalign{\smallskip}
\end{tabular}
\end{table}

As expected, with LM rescoring, the difference between the two models (baseline versus best) is smaller, compared to when the acoustic models are compared in their own right (first row of Table~\ref{tab:LM_Results}). Furthermore, we note that a baseline system tested on data which overlaps with the data on which the LM was obtained (`LM in favour') achieves a much better WER than the best model.

These results indicate the best performance we can expect when a language model is incorporated, both with the best augmented model, and with a baseline trained on a very limited (ca. 7h) training corpus. However, the use of an LM also diverts attention from the main question, which focuses on the most effective pretraining methods for acoustic models in low-resource scenarios. Thus, in what follows, all results are reported with no LM rescoring.


\section{Experiments}
\label{sec:experiments}
In this section, we turn to the experiments on data augmentation for Maltese ASR. We first describe an initial experiment to validate the architecture which will be used throughout the remainder of the experiments. Each subsequent set of experiments is characterised by a pretraining and a fine tuning stage. Since our emphasis is on data augmentation, it is the pretraining stage that mainly distinguishes the approaches described here. Our goal is to identify to what extent different data augmentation strategies work for an under-resourced language such as Maltese.

\subsection{State of the art reference}\label{subsec:reference}
Prior to explaining our baseline architectures - as a reference (but not a baseline that we optimise further in this work), we fine-tune a multilingual Wav2Vec\-2 model for Maltese ASR speech on the MASRI-Headset corpus, in order to gauge how far the simpler baselines we experiment with are from a more traditional transfer-learning setup for stronger models. The Wav2Vec\-2 model learns a cross-lingual speech representation (XLSR), pre-trained on up to 128 different languages~\cite{babu21}. 

In order to be applied for ASR tasks, this architecture needs to be fine-tuned using a supervised learning method. To achieve this, a fully connected layer is appended to a pre-trained model and the connectionist temporal classification (CTC) loss function is used to make the model learn a character to sound map. Three pre-trained models with varying parameter sizes were fine-tuned using a common hyperparameter set. The models are fine-tuned for 90 epochs, as an optimizer, AdamW is used, with a starting learning rate of $3e-4$. An effective batch size of 400 per epoch is used. However this is due to the gradient accumulation mechanism were the models weights and biases are only changed every 4 epochs. 

The results are reported in Table~\ref{tab:sota_wav2vec}. When fine-tuning an XLS-R 300M (300 million parameters) model with the MASRI-Headset corpus, we obtain a WER of 44.74\%. When fine-tuning the XLS-R 1B (1 billion parameters) model with the MASRI-Headset corpus, we obtain a WER of 30.94\%. When fine-tuning the XLS-R 2B (2 billion parameters) model with the MASRI-Headset corpus, we obtain a WER of 21.45\%. While there is a clear trend towards improved WER with larger models, these results clearly show
that low-resource languages with very small corpora available for fine-tuning are still far off from desirable results, even when fine-tuning on the largest ASR models available. Despite this, these results represent, at the time of writing, the current best ASR results for Maltese. 
The experiments reported below are intended as a study into the best strategy of augmentation for low-resource languages in order to improve results for Maltese (and other languages) in future work.

\begin{table}[ht]
\caption{Results for state of the art ASR systems for Maltese}
\label{tab:sota_wav2vec} 
\centering
\begin{tabular}{cccc}
\noalign{\smallskip}\hline\noalign{\smallskip}
\textbf{Train} & \textbf{Test} & \textbf{CER} & \textbf{WER} \\
\noalign{\smallskip}\hline\noalign{\smallskip}
\multicolumn{4}{c}{\textbf{XLS-R Wav2Vec2 300M}} \\
\noalign{\smallskip}\hline\noalign{\smallskip}
\begin{tabular}[c]{@{}c@{}}M-Headset v2\\ 6h19m\end{tabular} & \begin{tabular}[c]{@{}c@{}}M-Headset v2\\ 19m\end{tabular} & 44.74\% & 10.24\% \\
\noalign{\smallskip}\hline\noalign{\smallskip}
\multicolumn{4}{c}{\textbf{XLS-R Wav2Vec2 1B}} \\
\noalign{\smallskip}\hline\noalign{\smallskip}
\begin{tabular}[c]{@{}c@{}}M-Headset v2\\ 6h19m\end{tabular} & \begin{tabular}[c]{@{}c@{}}M-Headset v2\\ 19m\end{tabular} & 30.94\% & 8.34\% \\
\noalign{\smallskip}\hline\noalign{\smallskip}
\multicolumn{4}{c}{\textbf{XLS-R Wav2Vec2 2B}} \\
\noalign{\smallskip}\hline\noalign{\smallskip}
\begin{tabular}[c]{@{}c@{}}M-Headset v2\\ 6h19m\end{tabular} & \begin{tabular}[c]{@{}c@{}}M-Headset v2\\ 19m\end{tabular} & 21.45\% & 5.15\% \\
\noalign{\smallskip}\hline\noalign{\smallskip}
\end{tabular}
\end{table}

\subsection{Baselines}\label{subsec:validation}
As our baselines, we compare two variants of the Jasper architecture described in Section~\ref{subsec:arch}:
\begin{enumerate}
    \item Jasper 12 $\times$ 1 SEP: This consists of 12 blocks, each with 1 sub-block and uses separable convolutions.\footnote{See: \url{http://www.openslr.org/12}}. This architecture was originally described in the context of experiments on 100 hours of English data obtained from the LibriSpeech corpus. During training, the model is exposed to audio files sampled at a rate of 16 khz, with a maximum length of 16.7 seconds. Silence is trimmed at the beginning and end of each audio file, and transcriptions are normalised to expand abbreviations, convert numbers to letters, normalise consecutive whitespace and lowercase all text. This normalisation process was optimised for English.
    \item Jasper\_an4\_new: This is an adaptation of Jasper 12 $\times$ 1 to the set of languages and the amount of data available for our experiments. The architecture consists of $4$ blocks with $2$ repetitions of the 1D convolutional layer, making it smaller than Jasper 12 $\times$ 1 SEP. This is motivated by the empirical observation that a leaner architecture exhibits more stable behaviour (in terms of WER) with less training data, making it our architecture of choice. In this architecture, we apply a dropout rate of $0.2$ and place no restriction of length of audio files. Silence is not trimmed and no text normalisation is performed. 
\end{enumerate}

First, we establish that the two models achieve comparable performance after training on 100h of English data from LibriSpeech, then we compare them on the low-resource scenario of the present work, training them on both a small sample of English data from LibriSpeech and Maltese data from the MASRI-HEADSET corpus, with approximately seven hours of data in each case. The results are reported in Table~\ref{tab:architecture_test}.

\begin{table}[ht]
\caption{Results on varying training set sizes for the two architectures}
\label{tab:architecture_test} 
\centering
\begin{tabular}{cccc}
\noalign{\smallskip}\hline\noalign{\smallskip}
\textbf{Train} & \textbf{Test} & \textbf{Loss} & \textbf{WER} \\
\noalign{\smallskip}\hline\noalign{\smallskip}
\multicolumn{4}{c}{\textbf{Jasper 12$\times$1 SEP}} \\
\noalign{\smallskip}\hline\noalign{\smallskip}
\begin{tabular}[c]{@{}c@{}}LibriSpeech\\ train-clean-100\end{tabular} & \begin{tabular}[c]{@{}c@{}}LibriSpeech\\ dev-clean\end{tabular} & 44.37 & 29.54\% \\
\begin{tabular}[c]{@{}c@{}}M-Headset v1\\ 7h34m\end{tabular} & \begin{tabular}[c]{@{}c@{}}M-Headset v1\\ 31m\end{tabular} & 177.86 & 99.87\% \\
\begin{tabular}[c]{@{}c@{}}LibriSpeech\\ 7h34m\end{tabular} & \begin{tabular}[c]{@{}c@{}}LibriSpeech\\ 31m\end{tabular} & 190.50 & 70.06\% \\
\noalign{\smallskip}\hline\noalign{\smallskip}
\multicolumn{4}{c}{\textbf{Jasper\_an4\_new}} \\
\noalign{\smallskip}\hline\noalign{\smallskip}
\begin{tabular}[c]{@{}c@{}}LibriSpeech\\ train-clean-100\end{tabular} & \begin{tabular}[c]{@{}c@{}}LibriSpeech\\ dev-clean\end{tabular} & 37.67 & 29.21\% \\
\begin{tabular}[c]{@{}c@{}}LibriSpeech\\ 7h34m\end{tabular} & \begin{tabular}[c]{@{}c@{}}LibriSpeech\\ 31m\end{tabular} & 103.88 & 48.61\% \\
\begin{tabular}[c]{@{}c@{}}M-Headset v1\\ 7h34m\end{tabular} & \begin{tabular}[c]{@{}c@{}}M-Headset v1\\ 31m\end{tabular} & 69.28 & 65.61\% \\
\begin{tabular}[c]{@{}c@{}}M-Headset v2\\ 6h19m\end{tabular} & \begin{tabular}[c]{@{}c@{}}M-Headset v2\\ 19m\end{tabular} & 58.65 & 63.82\% \\
\noalign{\smallskip}\hline\noalign{\smallskip}
\end{tabular}
\end{table}

The two architectures achieve comparable results (WER $\sim 29\%$) on relatively large training sets of around 100h. However, with a much smaller training set of less than 10h, Jasper\_an4\_new outperforms the other. Though performance on Maltese is far from optimal, these experiments yield a baseline, consisting of a relatively small architecture with minimal data augmentation beyond the random noising strategy incorporated in SpecAugment.

Furthermore, we note that this starting position is not too far off from our reference XLS-R 300M model. This emphasises the importance of augmentation for low-resource language ASR further. In what follows, we adopt our simple architecture and consider several ways to improve the results through pretraining, rather than model complexity, in order to assess the various effects of augmentation alone.

\subsection{Unsupervised transcription}\label{subsec:unsup-cycles}

In our first set of experiments, we consider unsupervised (noisy) transcription as a technique for data augmentation. The main question we address is whether, given a baseline system trained on a very small dataset (see section~\ref{subsec:validation}), it is possible to generate noisy data for further pretraining and subsequent fine-tuning. To this end, we experiment with noisy transcription of data in Maltese (limited to around 12h), but also in other languages. In the latter case, we ask whether using a baseline Maltese ASR model to transcribe data from Arabic, Italian or English, can yield suitable pretraining data for further fine-tuning. The rationale for using non-target language data is that, in the presence of limited audio data for Maltese, pretraining using a baseline Maltese ASR system amounts to `reinterpreting' the non-target data using a representation that should bring it closer to the target. At the same time, our selection of data from non-target languages rests on observations about their typological proximity or historical relationship to Maltese (see section ~\ref{sec:intro2}).

The following datasets are used to create noisy pretraining data:
\begin{itemize}
    \item Maltese: 12h of audiobooks in Maltese (Section~\ref{subsec:mal-abooks})
    \item English: 100h hours portion of the LibriSpeech corpus (Section~\ref{subsec:librispeech})
    \item Arabic: 100h portion of the MGB-2 Corpus (Section~\ref{subsec:arab-data})
    \item Italian: 118h from the Italian portion of the M-AILABS Speech Dataset (Section~\ref{subsec:ita-data})
\end{itemize}

In addition, we reserve the manually transcribed MASRI-HEADSET v2 corpus for fine-tuning, reserving 250 utterances for testing (Section~\ref{subsec:masri-headset}).

Pretraining proceeds in cycles, each consisting of a pretraining phase and a fine-tuning stage. Each cycle uses the fine-tuned model from the previous cycle to regenerate the noisy transcriptions for the pretraining. Each experiment consists of three cycles, the optimum having been determined empirically: beyond 3 cycles, we begin to observe diminishing returns and oscillating WER values.

\begin{table}[ht]
\caption{Results for different pretraining datasets, over 3 cycles}
\label{tab:cycles}       
\centering
\setlength{\tabcolsep}{2pt}
\begin{tabular}{cccc|cccc}
\noalign{\smallskip}\hline\noalign{\smallskip}
\textbf{Data} & \textbf{Cycle} & \textbf{Loss} & \textbf{WER} & \textbf{Data} & \textbf{Cycle} & \textbf{Loss} & \textbf{WER} \\
\noalign{\smallskip}\hline\noalign{\smallskip}
\multirow{3}{*}{MT} & 1 & 55.82 & 61.64\%          & \multirow{3}{*}{AR}  & 1 & 64.67 & 60.26\%          \\
                         & 2 & 49.35 & 63.17\%          &         & 2 & 64.34 & 59.15\%          \\
                         & 3 & 58.95 & \textbf{60.95\%} &         & 3 & 60.06 & \textbf{58.15\%} \\
\noalign{\smallskip}\hline\noalign{\smallskip}
\multirow{3}{*}{EN} & 1 & 59.01 & 58.46\%          & \multirow{3}{*}{IT} & 1 & 57.36 & 60.22\%          \\
                         & 2 & 53.43 & 58.73\%          &         & 2 & 60.84 & \textbf{57.12\%} \\
                         & 3 & 58.30 & \textbf{58.27\%} &         & 3 & 59.99 & 63.06\%   \\   
\noalign{\smallskip}\hline\noalign{\smallskip}                    
\end{tabular}%
\end{table}


Table~\ref{tab:cycles} summarises the results on all experiments, reporting the final test-set WER for each language used for data augmentation, over three pretrain-and-fine-tune cycles. Two main conclusions can be drawn from these results. First, pretraining with around 100h of noisy data from non-target languages yields improved WER results. Second, the same noisy transcription strategy applied to a smaller amount of data from the target language itself also yields benefits (see the top panel in Table~\ref{tab:cycles}). However, at around 61\% WER, this is less than what we observe for pretraining over three cycles with English or Arabic. For both of these languages, we see a gradual WER drop over the three cycles. Similar observations apply to loss. On balance, with a lower loss and comparable WER, it appears that pretraining on English unsupervised (noisy) transcription is the more stable solution compared to Arabic. On the other hand, when pretraining on Italian we observe fluctuations in WER, with the best results obtained on the second cycle, though with a concomitant increase in loss.

Overall, the evidence suggests that non-target audio data in sufficient quantities, processed using a model trained on the target language, can yield modest improvements in target language ASR. Differential benefits are observed for pretraining in the non-target languages, perhaps in part due to the degree of lexical overlap between Maltese and the other three languages. A natural next step is to consider pretraining on data mixtures from different non-target languages, which also enables us to increase the pretraining dataset size. This is the focus of the next set of experiments.

\subsection{Language mixtures}\label{subsec:lang-mixes}
The rationale of the experiments in the previous subsection was that reinterpreting non-target data using a baseline ASR system for the target language should map the data to a representation that the target ASR system can benefit from with further training. Under this working assumption, we now consider the effect of mixing data from combinations of the three non-target languages, with a view to increasing the amount of pretraining data. Our goal is to determine if gains in terms of WER increase with dataset size when the data consists of language mixtures. 

We use the same pretraining data as in the previous subsection, presenting results for pretraining with combined datasets of two or more languages, including the small dataset of Maltese audiobook data. In what follows, we only report results for combinations involving English, alone or with other languages, based on the observation in the previous section that pretraining on English data results in stable gains in both WER and loss. Each experiment consists of a single cycle of pretraining followed by fine-tuning, following the same procedure as in Section \ref{subsec:unsup-cycles}.


\begin{table}[ht]
\caption{Results of mixed language pretraining using noisy transcription. Language columns indicate the amount of pretraining data in a given experiment.}
\label{tab:lang-mixes}       
\centering
\begin{tabular}{llllll}
\noalign{\smallskip}\hline\noalign{\smallskip}
\textbf{EN} & \textbf{MT} & \textbf{IT} & \textbf{AR} & \textbf{Loss} & \textbf{WER}\\
\noalign{\smallskip}\hline\noalign{\smallskip}
100h & 12h24m & - & - & $55.85$ & $59.84\%$ \\
100h & - & 118h41m & - & $56.90$ & $58.96\%$ \\
100h & - & - & 100h &  $61.33$ & $58.38\%$ \\
100h & - & 118h41m & 100h & $63.98$ & $59.15\%$ \\
100h & 12h24m & - &  100h & $58.95$ & $59.04\%$ \\
100h & 12h24m & 118h41m & 100h & $63.71$ & $57.96\%$ \\
\noalign{\smallskip}\hline
\end{tabular}
\end{table}

Table~\ref{tab:lang-mixes} displays results for different language mixtures. The results suggest that the benefits of this strategy are limited, and no notable gains in WER accrue with an increase in pretraining data; nor are there notable improvements from mixing languages. However, two further observations are worth making. First, mixing a relatively large corpus of English with a small sample of Maltese, the target language, yields WER improvements that outweigh those observed from using Maltese alone (see Table~\ref{tab:cycles}). Second, the best results are obtained when pretraining on all four languages; however, the benefits are modest compared to the results observed in the previous section with individual languages.


\subsection{Pretraining using supervised transcriptions and noisy annotation}
\label{subsec:sup-unsup}
Finally, we alter the pretraining strategy, while maintaining the same rationale of augmenting data using language mixtures from related languages. We noted above that mixing English and Maltese in the pretraining yields better WER scores than only using a limited amount of Maltese (compare Table \ref{tab:lang-mixes} and Table \ref{tab:cycles}). Rather than pretrain using English data transcribed with a baseline Maltese ASR system, we now use the gold English annotations in the LibriSpeech corpus, and mix these with noisy transcriptions in Maltese during pretraining.

\begin{table}[ht]
\caption{Results of mixed language pretraining using gold annotations for English}
\label{tab:sub-and-unsup}       
\centering
\begin{tabular}{llll}
\hline\noalign{\smallskip}
\textbf{EN} & \textbf{MT} & \textbf{Loss} & \textbf{WER}  \\
\noalign{\smallskip}\hline\noalign{\smallskip}
100h (gold) & 12h24m & $48.62$ & $55.05\%$ \\
\noalign{\smallskip}\hline
\end{tabular}
\vspace{-2mm}
\end{table}

Table~\ref{tab:sub-and-unsup} shows that this strategy yields a WER of around 55\%. This exceeds the best results obtained for the language mixture experiment. It shows the benefits of using supervised transcriptions for a language that is related to the target language in addition to noisy unsupervised annotations in the target language.

The experiments in this subsection suggest that not much is gained from mixing languages during pretraining using noisy data, compared to pretraining on individual languages. On the other hand, surprising improvements are obtained from mixing gold English transcriptions with noisy transcription of a small dataset in the target language. In the following section, we examine to what extent increasing the pretraining data in the target language using artificial means is beneficial on its own, as well as in tandem with gold data in English.

\subsection{Synthezising training data}
\label{subsec:synthetic}
In this section, we describe experiments using speech synthesis as a data augmentation tool, using the synthesized corpus described in Section \ref{subsec-masri-synth}. Despite the fact that the synthesis method yields noisy, imperfect data, synthetic training data has the advantage of being in potentially unlimited supply. In line with the conclusions from the previous subsection, we also consider the benefits of combining synthetically augmented target language data with gold data in English. In this way, we compare the trade-off between increasing data in the target language using artificial means (synthesis) versus adding data from a related language. We use the Maltese synthesized speech corpus described in Section~\ref{sec:data}, as well as the gold transcriptions from the 100h of LibriSpeech English data used previously.

We present results from two experiments. First, we establish baseline results using only the MASRI-SYNTHETIC Corpus. Next, we extend the findings from the previous subsection, where we find that the inclusion of gold English transcribed data with noisy Maltese data yields the best improvements. Here, we examine the impact of mixing gold English transcribed data with the synthesized data during pretraining. In both cases, as before, we fine-tune on the MASRI-HEADSET v.2 dataset.

\begin{table}[ht]
\caption{Results of augmentation using synthesised speech}
\label{tab:synth-speech}       
\centering
\begin{tabular}{lll}
\noalign{\smallskip}\hline\noalign{\smallskip}
\textbf{Pretraining} & \textbf{Loss} & \textbf{WER} \\
\noalign{\smallskip}\hline\noalign{\smallskip}
MT (synth) & 48.63 & 52.03\% \\
MT (synth) + EN 100h (gold) & 42.78 & 48.97\% \\
\noalign{\smallskip}\hline
\end{tabular}
\end{table}

Results are presented in Table~\ref{tab:synth-speech}. The use of 100h of synthesized speech brings our best WER down to 52\%, despite the noise present in the data. Once again, we observe a benefit from the inclusion of gold annotations in English during pretraining, with WER dropping to just under 49\%, which is an absolute reduction of 15\% in WER from our baseline systems (63.71\% WER).

\subsection{Summary of experiments and findings}
The experiments described in this section addressed different methods for data augmentation to overcome bottlenecks in the training data for an under-resourced language. Broadly speaking, we can characterise the rationale in terms of two distinct strategies: (a) pretraining by combining limited target data with comparatively large datasets from languages which are typologically or historically related to the target language; (b) pretraining using synthesized data. An important variable was the extent to which the data used in pretraining was `noisy': in the case of pretraining with data from other languages, we experiment with both a noisy transcription using a baseline ASR system, and gold annotations. In the case of synthesis, we expect the data to be imperfect in view of the limitations of the available TTS system.

Our experiments suggests that language mixtures yield benefits, even when the transcription is noisy. However, gold annotations are better; in particular, the inclusion of English gold data with noisy Maltese transcriptions yielded significant gains. Furthermore, we observe that pretraining on imperfect synthesized data in the target language also improves performance, with further gains provided once more by the inclusion of gold English data: an absolute reduction of 15\% in WER from the baseline system.


\section{Conclusions and future work}
\label{sec:conclusions}
The work presented in this paper had one over-arching theme - the analysis of whether ASR performance for under-resourced languages can be improved by various data augmentation methods. We elicited a detailed taxonomy of what types of augmentation exist in the literature, and we also devised a consistent set of experiments that sift through the many possibilities, whilst providing clear results for each. The results obtained are specific to ASR for the Maltese language, but we postulate that similar approaches would yield similar improvements for other under-resourced languages.

In order to assess that we were indeed testing for improvements in acoustic modelling based on data augmentation, this work specifically makes no use of language modelling as part of the ASR pipeline. We do however show that language model rescoring does indeed have a big effect on WER (section~\ref{subsec:no-lm}), and in future experimentation, we intend to combine the best methods from this work with further language modelling experiments to obtain state-of-the-art performance for Maltese ASR. That is however, a different research question altogether.

As part of our efforts in maintaining consistency in our experiments, questions were also raised as to whether augmentation is helpful for pretraining methodologies, whilst keeping the network architectures unchanged. To this end we employ training cycles, and show that some measure of performance improvement is obtained, as shown in Section~\ref{subsec:unsup-cycles}. We also postulate that this approach is generic enough to be proposed for any ASR setup for an under-resourced target language.

The one aspect of the work we present here that is probably dependent on the target language is in section~\ref{subsec:lang-mixes}. The ancillary language data chosen is not random. This augmentation is based on languages which are somehow related to Maltese - either historically, culturally or linguistically. The results show that supervised transcriptions of some closely related languages provide a substantial improvement in WER (Section~\ref{subsec:sup-unsup}). The choice of these languages, however, has to be assessed individually based on the target language.

Furthermore, this paper then analyzed the use of synthesized speech for ASR training (Section~\ref{subsec:synthetic}). Whilst data from a speech synthesis system might not always be readily available for under-resourced languages, we highlight the fact that the synthetic speech quality need not be of very good quality for immediate benefits in WER. In fact, the synthesis system used is a minimalistic diphone-based system (which can be built for any low-resource language) produces highly unnatural speech with many pronunciation errors, yet we gained the largest improvements from using this method of data augmentation.

In summary, the conclusions we can draw from the work presented is as follows:
\begin{itemize}
    \item Both gold and noisy transcriptions can be used as data augmentation techniques, up until a limit is reached after various training cycles.
    
    \item Noisy transcriptions have poorer performance with respect to gold transcriptions. In the absence of sufficient gold transcription quantities, however, the utilization of noisy transcriptions of the target language shows substantial improvements in WER.
    
    \item Mixing pretraining data from non-target languages is useful, especially with a small data batch from the target language. There are however limits to this and adding more data beyond a certain point yields diminishing returns.
    
    \item The use of synthesized speech in the target language, as training data, outperforms the use of gold or noisy transcriptions in languages different to the target, even when the synthesized speech has a low quality.
\end{itemize}

The results we obtain from this work are very encouraging, as we have observed an absolute reduction of 15\% in WER from our baseline systems (63.71\% WER) to the system resulting from the best setup based on pretraining and fine tuning from an augmented dataset (48.97\% WER), approaching the result of a 300 million parameter XLSR Wav2Vec2 fine-tuned model. This is a remarkable improvement in light of the fact that they revolve around a curated Maltese ASR corpus of less than $7$ hours of speech. In the interim, we have also started a crowd-sourced data collection effort on the Mozilla CommonVoice platform, which has collected a few more hours of labelled Maltese speech data for ASR use.

We believe this work sets a starting point for all future work in ASR for the Maltese language, and propose that all languages which are similarly under-resourced could follow the methodology we outline in this paper to assess and capitalize on data augmentation methods, when no other possibilities are present.

As indicated earlier in the paper, the most promising direction for future work is to explore the augmentation techniques presented here with a different neural architecture. Recent research in multilingual NLP has shown the remarkable performance of models in the Transformer paradigm~\cite{Vaswani2017}, both for textual, language-understanding tasks with models such as multilingual BERT~\cite{Devlin2019}, XLSR and XLS-R~\cite{Conneau2020,babu21} and for speech recognition with Wav2Vec-2~\cite{baevski2020wav2vec}. These approaches provide impetus for further experimentation, with a view to identifying the best strategies for data augmentation to make ASR more feasible for low-resource languages.

\bibliographystyle{IEEEtran}
\bibliography{mybib}


\begin{IEEEbiographynophoto}{Andrea DeMarco}
read for his undergraduate degree in Information Technology (majoring in Computer Science and Artificial Intelligence) from the University of Malta, which he followed with an M.Sc. in Computer Science and Artificial Intelligence from the same institution. Following this, he moved to the UK where he completed a Ph.D. at the Audio, Speech and Language Processing Laboratory of University of East Anglia, one of the major centers of speech processing. His area of research was in accent classification and modeling from speech using acoustic methods and machine learning. Following his PhD, he joined the Institute of Space Sciences and Astronomy (ISSA) at the University of Malta as a post-doctoral researcher involved in the construction of various sub-elements of the Square Kilometre Array (SKA) radio telescope. Following this stint, he was promoted with a lectureship within the institute. As of January 2019, he has started a collaboration with the Institute of Linguistics and Language Technology at the University of Malta, as an Affiliate member, focusing on speech technology research, in particular for the Maltese language.
\end{IEEEbiographynophoto}

\begin{IEEEbiographynophoto}{Carlos Mena}
read for his degree in Communications and Electronics Engineering at the Instituto Politécnico Nacional (IPN) in Mexico City, then he followed with a M. Eng. in Computer Engineering at the Universidad Nacional Autónoma de México (UNAM) where he subsequently completed his Ph.D. in Digital Signal Processing, focusing on Automatic Speech Recognition. He was a post-doc at the Institute of Linguistics and Language Technology (ILLT) from the University of Malta. In 2021, he moved to the University of Reykjavík as a post-doc of the Language and Voice Laboratory (LVL). His main interests are Automatic Speech Recognition for minority languages, corpus curation and creation and applied phonetics.
\end{IEEEbiographynophoto}

\begin{IEEEbiographynophoto}{Albert Gatt}
read for his undergraduate degree in Psychology and Linguistics at the University of Malta, where he subsequently obtained a M.A. in Linguistics (formal semantics). He completed his Ph.D. at the University of Aberdeen, UK, in Computing Science, focusing on the automatic generation of text from non-linguistic data. He was a post-doc at the Department of Computing Science, University of Aberdeen until 2009, when he moved to the University of Malta, where he was Director of the Institute of Linguistics and Language Technology for eight years. In 2021, he moved to the Department of Information and Computing Sciences at Utrecht University in the Netherlands, where he is a member of the NLP group. His main research interests are in Natural Language Generation, NLP for under-resourced languages, and deep neural models for multimodal language processing, especially those incorporating vision and speech.
\end{IEEEbiographynophoto}

\begin{IEEEbiographynophoto}{Claudia Borg}
received her B.Sc., M.Sc., and Ph.D. degrees from the University of Malta in 2005, 2009 and 2016 respectively. Her research is focused on the processing of the Maltese Language, with a particular interest in computational morphology for Maltese as a low-resource setting. She is currently a Senior Lecturer with the Department of Artificial Intelligence at the University of Malta. 
\end{IEEEbiographynophoto}

\begin{IEEEbiographynophoto}{Aiden Williams}
read for his undergraduate degree in Artificial Intelligence at the University of Malta and is set to graduate in November 2022. His resesrch focus was the evaluation of state-of-the-art ASR systems for low-resource languages, specifically Maltese.
\end{IEEEbiographynophoto}

\begin{IEEEbiographynophoto}{Lonneke van der Plas}
obtained her undergraduate degree from the Radboud University and an M.Phil in Computer Speech and Language Processing from the University of Cambridge. Her PhD is from the University of Groningen on the topic of computational semantics and question answering systems. She did a post-doc at the University of Geneva and subsequently held a junior professorship at the University of Stuttgart until she moved to the University of Malta in 2014. Since 2021, she is leading a research group at the Idiap research institute in Martigny, Switzerland. Her main research interests are in computational semantics, cross-lingual NLP also for under-resourced languages, and computational creativity.
\end{IEEEbiographynophoto}

\vfill

\end{document}